\documentclass[sigconf]{acmart}
\usepackage{algorithm}
\usepackage{algpseudocode}
\usepackage{amsmath}
\usepackage{subfigure}
\usepackage{caption}
\usepackage{multirow}
\usepackage{diagbox}
\usepackage{balance}
\AtBeginDocument{%
  \providecommand\BibTeX{{%
    \normalfont B\kern-0.5em{\scshape i\kern-0.25em b}\kern-0.8em\TeX}}}

\copyrightyear{2021}
\acmYear{2021}
\setcopyright{acmcopyright}\acmConference[MM '21]{Proceedings of the 29th ACM International Conference on Multimedia}{October 20--24, 2021}{Virtual Event, China}
\acmBooktitle{Proceedings of the 29th ACM International Conference on Multimedia (MM '21), October 20--24, 2021, Virtual Event, China}
\acmPrice{15.00}
\acmDOI{10.1145/3474085.3475544}
\acmISBN{978-1-4503-8651-7/21/10}
\begin{document}

\fancyhead{}
\title{SimulSLT: End-to-End Simultaneous Sign Language Translation}

\author{Aoxiong Yin$^{1}$, Zhou Zhao$^1$*, Jinglin Liu$^{1}$, Weike Jin$^{1}$,  Meng Zhang$^2$,  Xingshan Zeng$^2$,  Xiaofei He$^1$}

\makeatletter
\def\authornotetext#1{
\if@ACM@anonymous\else
    \g@addto@macro\@authornotes{
    \stepcounter{footnote}\footnotetext{#1}}
\fi}
\makeatother
\authornotetext{Corresponding author.}

\affiliation{
 \institution{\textsuperscript{\rm 1}Zhejiang University}
 \institution{\textsuperscript{\rm 2}Huawei Noah's Ark Lab}
 }
\email{ {yinaoxiong,zhaozhou,jinglinliu, weikejin}@zju.edu.cn, zhangmeng92@huawei.com, zxshamson@gmail.com, xiaofei_h@qq.com}

\def\authors{Aoxiong Yin, Zhou Zhao, Jinglin Liu, Weike Jin, Meng Zhang, Xingshan Zeng, Xiaofei He}
\renewcommand{\shortauthors}{Yin and Zhao, et al.}

\begin{abstract}
  Sign language translation as a kind of technology with profound social significance has attracted growing researchers' interest in recent years. However, the existing sign language translation methods need to read all the videos before starting the translation, which leads to a high inference latency and also limits their application in real-life scenarios. To solve this problem, we propose SimulSLT, the first end-to-end simultaneous sign language translation model, which can translate sign language videos into target text concurrently. SimulSLT is composed of a text decoder, a boundary predictor, and a masked encoder. We 1) use the wait-k strategy for simultaneous translation. 2) design a novel boundary predictor based on the integrate-and-fire module to output the gloss boundary, which is used to model the correspondence between the sign language video and the gloss. 3) propose an innovative re-encode method to help the model obtain more abundant contextual information, which allows the existing video features to interact fully. The experimental results conducted on the RWTH-PHOENIX-Weather 2014T dataset show that SimulSLT achieves BLEU scores that exceed the latest end-to-end non-simultaneous sign language translation model while maintaining low latency, which proves the effectiveness of our method.
\end{abstract}




\maketitle

\section{INTRODUCTION}
Sign language is a visual language widely used by an estimated 466 million deaf or hard-of-hearing people \cite{web1}. They use various methods to convey information, such as gestures, movements, mouth shapes, facial expressions, etc \cite{DBLP:conf/ejc/Medvedev08}. However, there are certain obstacles for normal people to understand this language, which causes inconvenience to their daily lives. This motivates researchers to design the AI sign language translation system \cite{DBLP:conf/icimcs/ZhangZL14,DBLP:journals/ijcv/KollerZNB18,DBLP:conf/cvpr/KollerZN17}.

\begin{figure}[tb]
  \includegraphics[width=\linewidth]{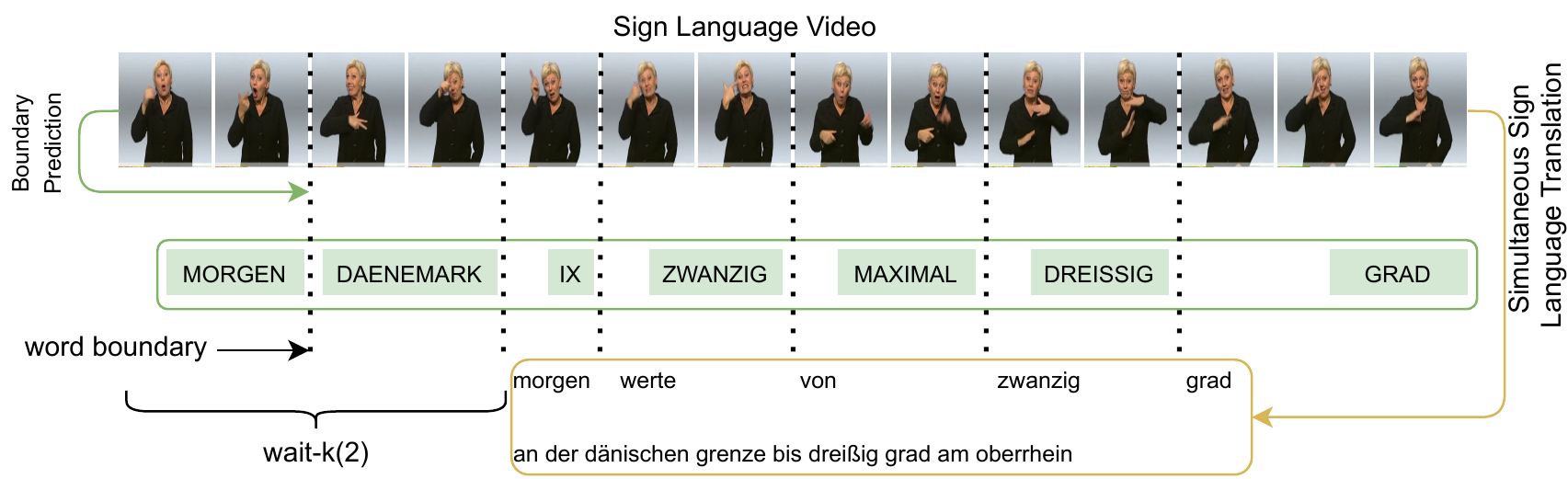}
  \caption{An example to illustrate how our model works. The text in the green box represents sign language gloss, and the text in the yellow box represents spoken language.}
  \label{fig:ex}
\end{figure}

Previous studies on sign language translation (SLT) mainly focus on non-simultaneous translation methods \cite{DBLP:conf/cvpr/CamgozKHB20,dataset3,DBLP:conf/eccv/CamgozKHB20,DBLP:conf/nips/LiX0ZSSL20,Yin2020,Ko2019,DBLP:conf/aaai/GuoZLW18}. The model needs to read an entire video before starting translation. Therefore, if they are applied to simultaneous scenes in actual use, such as assisting deaf or hard-of-hearing people to seek help from staff in public places like airports, they will all suffer high inference delays without exception. This is unacceptable for practical applications. 

Therefore, we need a simultaneous sign language translation system that can better balance translation quality and delay to meet the needs of practical applications. Simultaneous sign language translation is a very challenging task. For example in the simultaneous translation task, the model needs to decide when to switch from the encoding state to the decoding state. In simultaneous text translation, this decision is made at the token (word or BPE) level \cite{DBLP:conf/ijcnlp/MaPK20}. However, in this task, since there is no clear semantic unit and its corresponding boundary, we need to design a suitable boundary predictor, to make this decision. 

Hence, we propose the SimulSLT model, an end-to-end simultaneous sign language translation model based on Transformer. 
The main model is mainly composed of three parts, 1) a masked encoder for encoding streaming input sign language video; 2) a text decoder where the cross-attention follows the wait-k strategy \cite{Ma2019}; 3) a boundary predictor for predicting the boundary of gloss words. An example to illustrate how our model works is shown in Figure \ref{fig:ex}. First, the sign language video is divided into segments corresponding to gloss by the boundary predictor. Next, after waiting for the video frames corresponding to k (here is 2) glosses, the model starts to translate the target text, and then translates a text every time it encounters a word boundary. Until all the videos have been read, the model enters the non-simultaneous translation state. To get the video segment corresponding to the gloss, we design a novel boundary predictor based on the integrate-and-fire mechanism, which works similarly to human neurons. The stimulus signal generated by the streamed sign language video is continuously accumulated in the boundary predictor until the signal exceeds the threshold value of the boundary predictor to fire the word boundary, then the signal falls back, and the boundary predictor enters the next round of accumulation. 

However, due to the lack of available alignment annotation information, we cannot directly train the boundary predictor. To solve this problem, we introduce an auxiliary gloss decoder to help the boundary predictor learn alignment information.

In terms of context feature acquisition, most of the previous work directly used the output of the encoder before the boundary predictor as the context feature for text decoding. In this case, each video frame can only interact with the video before it, which makes the interaction between known video information during decoding insufficient. Therefore, we design an innovative re-encode method to let the output of the encoder contain more abundant contextual information, which can make the obtained video frame information fully interact with each other while ensuring that the algorithm time complexity is kept at \(O(n^2)\). 

Considering the difficulty of simultaneous sign language translation tasks, we introduce a knowledge distillation method to help model optimization, which is used to transfer knowledge from non-simultaneous teacher models to SimulSLT. In addition, we also add an auxiliary connectionist temporal classification (CTC) decoder behind the encoder to enhance the feature extraction capability of the encoder and help the boundary predictor to better learn alignment information. The experimental results show that on the RWTH-PHOENIX-Weather 2014T (PHOENIX14T) dataset \cite{dataset3}, SimulSLT even obtains BLEU scores that exceed the latest end-to-end non-simultaneous sign language translation model while maintaining low latency, which proves the effectiveness of our method. We also conduct ablation experiments to verify the effectiveness of all proposed methods in SimulSLT.

To summarize, the contributions of this work are as follows:
\begin{itemize}
  \item We are the first to explore the simultaneous sign language translation problem, and we propose SimulSLT, which is a simultaneous sign language translation model based on wait-k strategy.
  \item We design a novel boundary predictor to help the model decide when to switch from the encoding state to the decoding state. It efficiently learns the correspondence between sign language video and gloss. 
  \item We design a general re-encode to help the simultaneous translation model to obtain more abundant contextual information.
  \item Extensive experiments are conducted on the public benchmark dataset (PHOENIX14T) and the results show that SimulSLT achieves translation accuracy higher than the latest end-to-end non-simultaneous sign language translation methods while maintaining low latency. A broad range of new baseline results can guide future research in this field.
\end{itemize}

\section{RELATED WORKS}
Research into sign language translation has a long history \cite{old1,old2,old3,JinTao,WencanHuang}. In recent years, with the rise of deep learning, many people have tried to use neural network methods to deal with SLT tasks and achieved good results. Due to lack of data, early research mainly focused on Isolated Sign Language Recognition \cite{islr1,islr2,islr3,islr4,islr5,islr6}. In recent years, with the emergence of a series of high-quality datasets \cite{dataset1,dataset2,dataset3}, researchers have begun to turn to the study of continuous sign language recognition (CSLR) and sign language translation (SLT). The CSLR task aims to convert the sign language video into the corresponding sign language gloss, and these two sequences have the same order. However, the equivalent spoken language that can be understood by ordinary people is different in length and order compared with gloss. There is a more detailed introduction about the difference between CSLR and SLT in \cite{dataset3}.

Sign language translation aims to translate a continuous sign language video into the corresponding spoken language. \citet{dataset3} first formalize the sign language translation task in the framework of neural machine translation (NMT) and released the first publicly available sign language translation dataset, RWTH-PHOENIX-Weather-2014T (PHOENIX14T). Soon after, They use the Transformer structure to design an end-to-end translation model using gloss and text as supervision signals, instead of relying on the CSLR model and using gloss as an intermediate representation \cite{DBLP:conf/cvpr/CamgozKHB20}. This work shows that using gloss as an intermediate language is not a good choice. Compared with the two-stage model, the end-to-end video to text model can better perform model training and information transmission. The work of \citet{Yin2020} further proves this point. They obtain better performance by using a STMC network for tokenization instead of translating GT glosses.

The previous work has used gloss annotations as auxiliary supervision signals \cite{Ko2019,Yin2020,DBLP:conf/cvpr/CamgozKHB20,dataset3,DBLP:conf/icip/SongGXW19}. However, the acquisition of gloss annotations is expensive and difficult since it requires sign language experts to annotate. Therefore, it is necessary to explore how to train the model when the gloss label is missing. \citet{DBLP:conf/nips/LiX0ZSSL20} reduced the model's dependence on gloss annotation by fine-tuning the feature extractor on the word-level sign language dataset of another different sign language. This shows that different sign languages have a lot in common in the underlying features. The work of \citet{Orbay2020} showed that good hand shape representations could improve translation performance, which is consistent with our common sense because sign language usually conveys much information through gestures.

\begin{figure*}[tbp]
  \setlength{\abovecaptionskip}{2pt}
  \centering
  \subfigure[The overview of the model architecture for SimulSLT]{
    \begin{minipage}[t]{0.55\linewidth}
      \centering
      \includegraphics[width=\linewidth]{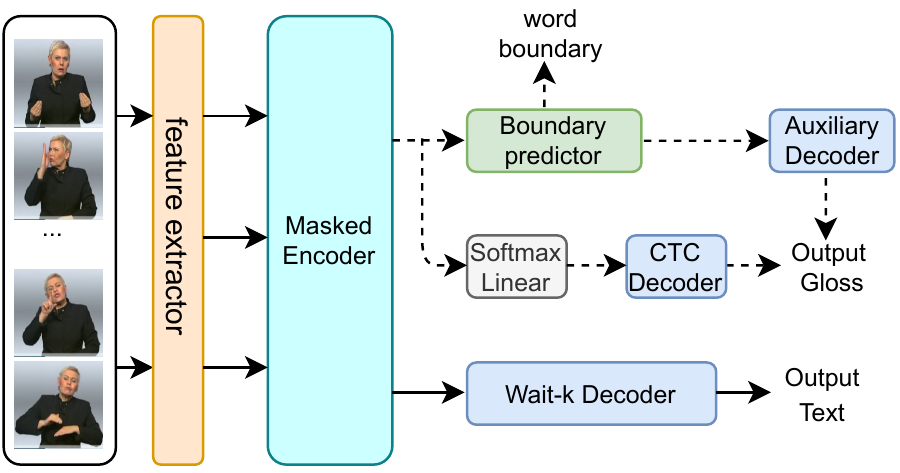}
      \label{fig:main-a}
    \end{minipage}
  }%
  \subfigure[Detailed structure and workflow of the masked encoder]{
    \begin{minipage}[t]{0.45\linewidth}
      \centering
      \includegraphics[width=\linewidth]{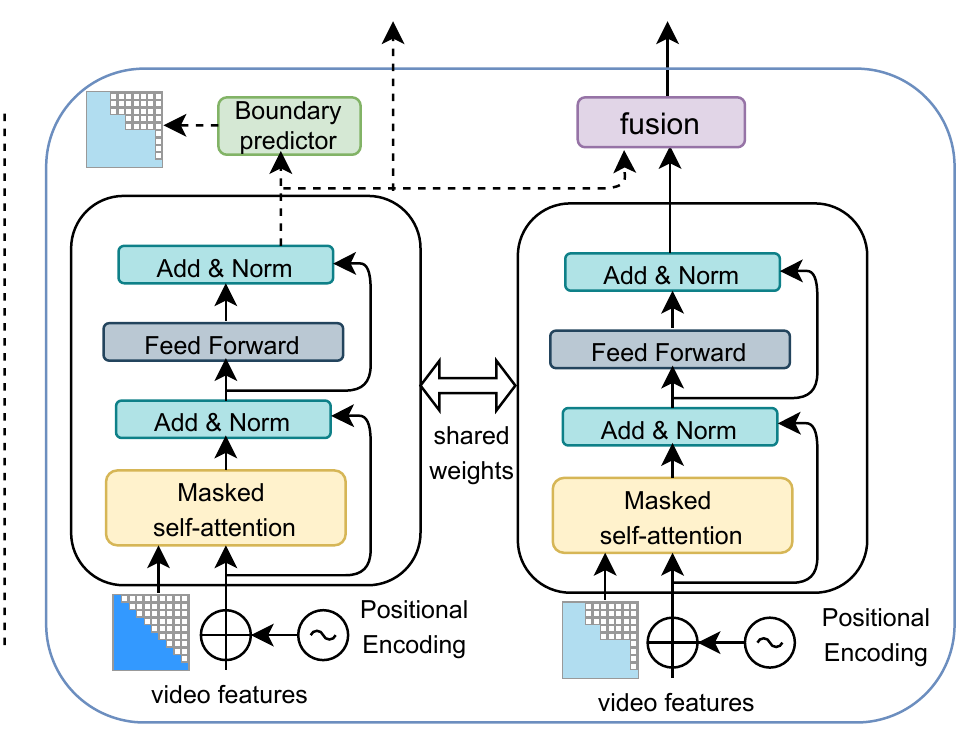}
      \label{fig:main-b}
    \end{minipage}
  }
  \centering
  \caption{(a) The model structure of SimulSLT. The dotted line represents the output of the encoder after the first encoding, and the solid line represents the output of the encoder after re-encode.  (b) Detailed structure and workflow of the masked encoder.}
  \label{fig:main}
\end{figure*}

Delay is also essential for sign language translation. However, the research that has been conducted in SLT to date have to read a complete sign language video to start the translation, which will cause a serious out of sync between the signer and the text generated by the model. Therefore, we are the first to explore the simultaneous sign language translation problem. Research on simultaneous translation in other fields has made some progress \cite{Ma2019,DBLP:conf/acl/ArivazhaganCMCY19,DBLP:conf/iclr/MaPCPG20}. \citet{Ma2019} present a very simple yet effective "wait-k" policy for simultaneous text translation model. The model will first wait k source words and then translates concurrently with the rest of source sentence. By applying "wait-k" policy to the field of simultaneous speech translation and combining multi-task learning and knowledge distillation methods, \citet{DBLP:conf/acl/RenLTZQZL20} proposed the SimulSpeech model. Their work proves that the end-to-end simultaneous translation system has lower latency and higher performance than the cascaded translation system.

The integrate-and-fire model is the most prominent spiking neuron model \cite{DBLP:journals/bc/Burkitt06,DBLP:journals/nn/Maass97}. It simulates the activity of neurons in biology. It increases the membrane potential according to the input of the neuron and fires after reaching the threshold. After firing, the membrane potential drops to a lower level and starts the next round of accumulation. It can model the relationship between continuous and discrete sequences very well. Recently, the integrate-and-fire module has been applied to speech recognition and lip recognition tasks and achieved good results \cite{DBLP:conf/icassp/Dong020,DBLP:conf/mm/LiuRZZHY20}.

\section{METHODS}
\subsection{Overview}
In this section, we introduce SimulSLT and describe our methods thoroughly. As shown in Figure \ref{fig:main-a}, the main model of SimulSLT is based on the Transformer \cite{DBLP:conf/nips/VaswaniSPUJGKP17} architecture and consists of a CNN video feature extractor, a masked encoder, a boundary predictor, and a wait-k decoder. In order to help the boundary predictor learn the alignment information better, we introduce an auxiliary gloss decoder and an auxiliary connectionist temporal classification (CTC) decoder. We designed a new re-encoding method to enhance the feature expression ability of the encoder and help the model obtain more abundant contextual semantic information. Besides, to further tackle the challenges in SimulSLT, we propose a knowledge distillation method to reduce the optimization difficulty of the model and improve the performance of the model. We will introduce them in detail in the following subsections.
\subsection{Masked Encoder}
As shown in Figure \ref{fig:main-b}, the encoder in SimulSLT consists of a stacked masked self-attention layer and a feedforward layer. In order to adapt to the task of simultaneous translation, we designed a masked self-attention mechanism to ensure that the current frame can only see the video frame before it. As shown in Figure \ref{fig:main-b}, the streaming sign language video features are first input to the encoder on the left, which is used to learn the alignment information between the video and the gloss, and provide the encoded features for the boundary predictor. We follow the boundary predictor with an auxiliary gloss decoder, which is used to provide alignment supervision information between the video and the gloss during training. A connectionist temporal classification (CTC) \cite{DBLP:conf/icml/GravesFGS06} decoder with CTC loss is also added to the encoder to learn more meaningful spatio-temporal representations of the sign. Whenever the wait-k strategy decides to switch to the decoding state and the video stream has not ended, as shown on the right in Figure \ref{fig:main-b}, we apply the re-encode method to re-encode the video features that have been obtained. This method will be introduced in detail in section \ref{section:re}. After that, we further merge the outputs of the two encoders and output them to the text decoder for decoding (here we use the direct addition method for fusion). The two encoders share the same weight. 

\subsection{Boundary Predictor}
We build our boundary predictor based on the Integrate-and-fire module. Like natural neurons, it emits pulses after signal accumulation reaches a threshold, and each pulse represents a word boundary. The encoder's output sequence \(H = (h_1,h_2,...,h_n)\) first passes through a multi-layer perceptron (MLP) to obtain the weight embedding sequence \(W = (w_1,w_2,...,w_n)\). The running process of the entire boundary predictor can be expressed by the following formula: 

\begin{gather}
  w=sigmoid((relu(hW_1+B_1)+h)W_2+B_2) \label{fm:MLP} \\
  b_j = \mathop{\arg\min}\limits_{t}(r_{j-1}+\displaystyle\sum_{i=b_{j-1}+1}^t{w_i}>T) \\
  r_j = r_{j-1}+\displaystyle\sum_{i=b_{j-1}+1}^{b_j}{w_i} - T
\end{gather}
\begin{gather}
  l_j = w_{b_{j}}-r_j \\
  e_j = r_{j-1}*h_{b_{j-1}} + \displaystyle\sum_{i=b_{j-1}+1}^{b_j-1}{w_i*h_i} + l_j*h_{b_j} \label{fm:gloss embedding}
\end{gather}
where formula \ref{fm:MLP} represents the calculation method of the MLP we designed, \(w\) represents the continuous stimulus signal in the time series, and a kind of technology module accumulates it and emits the word boundary \(b_j\) and gloss embedding \(g_i\) after reaching the threshold \(T\) (we set it to 1.0). When the accumulated stimulus signal reaches the threshold, the stimulus signal \(w_{b_j}\) will be split into two parts. The part one \(l_j\) within the threshold is used to calculate gloss embedding \(e_j\), and the remaining part \(r_j\) is used for the next accumulation to calculate \(e_{j+1}\). As shown in formula \ref{fm:gloss embedding}, gloss embedding is obtained by multiplying the weight within the threshold by encoder output and then adding them.


\subsection{Decoder}
\subsubsection{Auxiliary  Decoder}
The auxiliary decoder is a decoder made up of multiple layers of Transformer decoders \cite{DBLP:conf/nips/VaswaniSPUJGKP17} with the cross-attention mechanism removed because we already have an a kind of technology module to align source and target. The decoder takes in the gloss embedding sequence \(E = (e_1,e_2,...,e_n)\), and then generates the gloss tokens \(G = (g_1,g_2,...,g_n)\) . The decoding process only occurs during the training process for learning alignment information.

\subsubsection{Wait-k Decoder}
In the SimulSLT model, we adopt the wait-k strategy \cite{Ma2019} for simultaneous translation. Let \((x,y)\) be a pair of video-text sequences. Given the source and target context, the model need to compute the distribution over the next target token.

\begin{gather}
  P(y_t|y_{<t},x_{<t+k};\theta )
\end{gather}
where \(\theta \) is the model parameter, \(y_{<t}\) represents the target tokens before position \(t\), and \(x_{<t+k}\) represents the video segments before position \(t+k\) . To ensure that the decoder can only see \(t+k-1\) video segments when generating the target token \(y_t\), we generate a mask based on the word boundaries obtained by the previous prediction to mask the encoder's output when performing cross-attention.

\begin{figure}[htbp]
  \setlength{\abovecaptionskip}{2pt}
  \centering
  \subfigure[Not Re-encode]{
    \begin{minipage}[t]{0.3\linewidth}
      \centering
      \includegraphics[width=\linewidth]{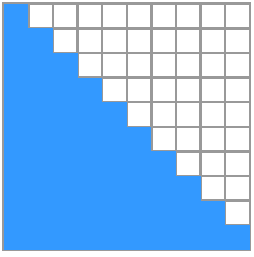}
      \label{fig:re-encode1}
    \end{minipage}
  }%
  \subfigure[Re-encode Once]{
    \begin{minipage}[t]{0.3\linewidth}
      \centering
      \includegraphics[width=\linewidth]{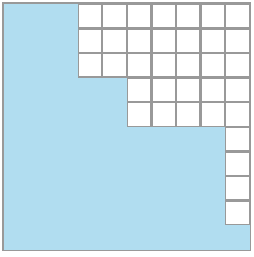}
      \label{fig:re-encode2}
    \end{minipage}
  }
  \subfigure[Re-encode Every Time]{
    \begin{minipage}[t]{0.3\linewidth}
      \centering
      \includegraphics[width=\linewidth]{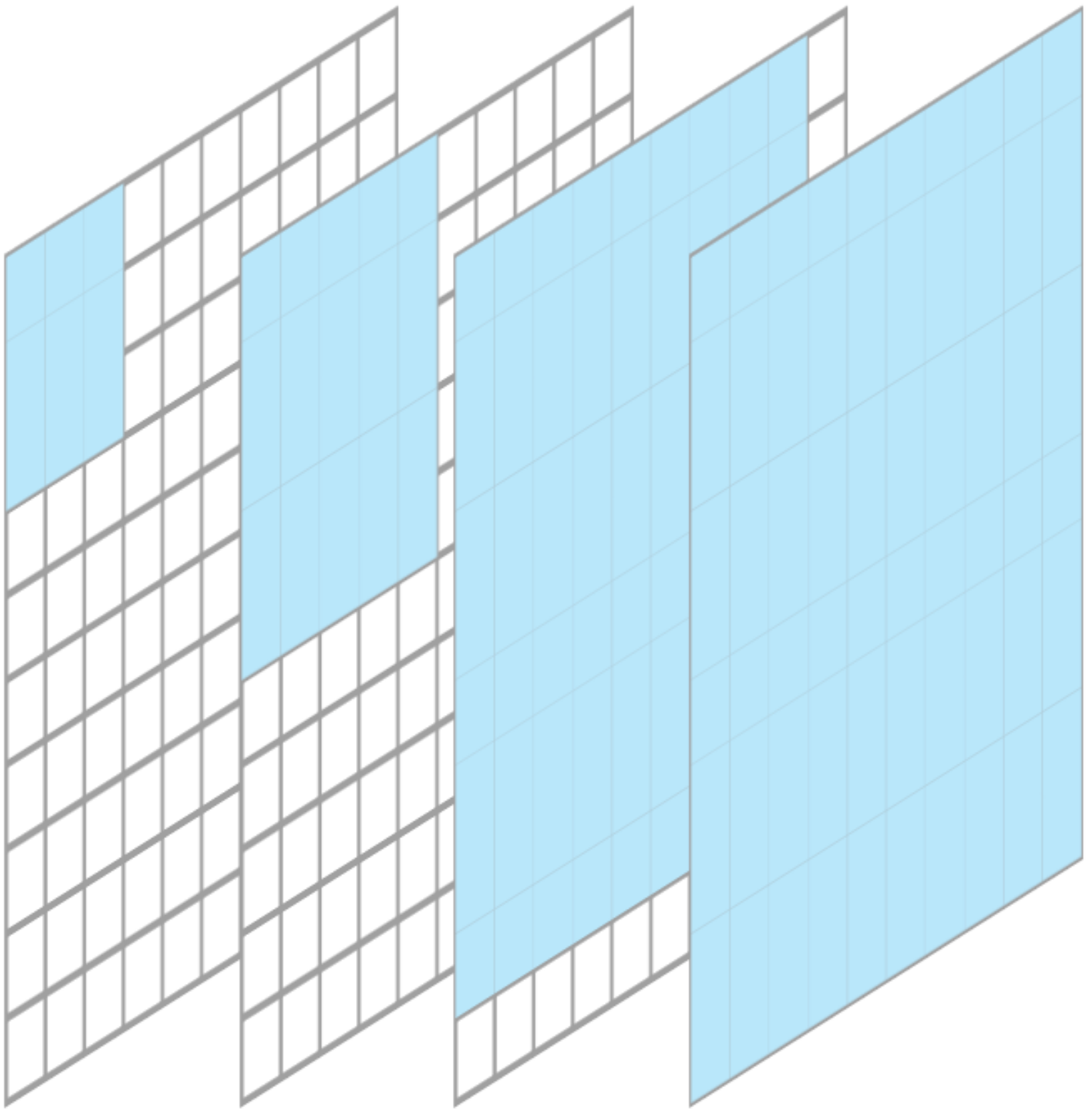}
      \label{fig:re-encode3}
    \end{minipage}
  }
  \centering
  \caption{An example to illustrate three different re-encode strategies.}
\end{figure}

\subsection{Re-encode Strategy} \label{section:re}
To simulate the real simultaneous translation task, most of the previous work used the lower triangular mask as shown in Figure \ref{fig:re-encode1} to ensure that the current frame can only see the frame before it, which is reasonable when performing word boundary prediction. However, directly using the encoder's output in the boundary prediction stage to decode the text will bring some performance loss because each frame only performs self-attention calculation with the frame before it, and there is no sufficient interaction between the known frames. 

An ideal re-encode method is shown in Figure \ref{fig:re-encode3}. Whenever we predict a word boundary, we re-encode all the previous frames until the end of the sequence. The calculation of the self-attention part of this algorithm is as follows:

\begin{gather}
  \begin{split}
  \hat{v}_j=\sum_{k}^{b_i}{\beta_kv_k},\ &where \  \beta_k=\frac{exp(sim(v_j,v_k))}{\sum_{q}^{b_i}exp(sim(v,v_q))} \\
  &j\leq b_i
  \end{split}
\end{gather}
where \(V=(v_1,v_2,...,v_n)\) represents the input sequence of self-attention, \(b_i\) represents the position of the i-th word boundary, \(\hat{v}_j\) represents the value obtained after self-attention, and the sim function uses the dot product function after the dimension scaling mentioned in \cite{DBLP:conf/nips/VaswaniSPUJGKP17}. Let us consider the worst case that this algorithm can encounter: a word boundary is predicted for each frame. In this case, its calculation time is as follows: 

\begin{gather}
  1^2+2^2+...+n^2=\frac{n(n+1)(2n+1)}{6} 
\end{gather}

Obviously the time complexity of the algorithm is \(O(n^3)\), while the time complexity of the original algorithm is \(O(n^2)\). At the same time, the algorithm requires re-encoding every time, which is not conducive to parallel training of the model. So we proposed an algorithm that only re-encode once as shown in Figure \ref{fig:re-encode2}. The self-attention calculation method of the improved algorithm is as follows:

\begin{gather}
  \begin{split}
    \hat{v}_j=\sum_{k}^{b_i}{\beta_kv_k},\ &where \  \beta_k=\frac{exp(sim(v_j,v_k))}{\sum_{q}^{b_i}exp(sim(v,v_q))} \\
    &b_{i-1}<j\leq b_i
  \end{split}
\end{gather}

In this algorithm, we only re-encode the frames in a word range, and the frames in the previous word range no longer calculate self-attention with the following frames. In this way, we can reduce the algorithm's time complexity to \(O(n^2)\) without affecting the expression of word information.

In addition, since it only needs to re-encode once to get all the output needed for decoding, this allows us to train the model efficiently and in parallel. Our subsequent experimental results prove that this recoding strategy helps to improve the performance of the model. 


\subsection{Knowledge Distillation}

Knowledge distillation \cite{DBLP:journals/corr/HintonVD15,kim-rush-2016-sequence} is widely used to reduce the model's delay and improve the performance of the student model. In our work, we use this method to transfer knowledge from a non-simultaneous teacher model to a SimulSLT model. First, we input the source video \(x\) into the teacher model to get logit \(z\). Then input \(z\) into the softmax-T function, which is used to construct a soft target \(y'\) to obtain information from the negative label.

\begin{gather}
  y'_i=\frac{exp(z_i/\Gamma )}{\sum_{j} exp(z_j/\Gamma )  } 
\end{gather}

where \(y'\) the soft target which is used to supervise the training of the student model.

\subsection{Training Of SimulSLT}
In The SimulSLT model, the CTC decoder is used to enhance the expressive ability of the encoder, and we use CTC loss to optimize it. CTC introduces a set of intermediate paths \(\phi (y)\) called CTC path for a target text qequence \(y\). Multiple CTC paths may correspond to the same target text sequence, because the length of the source sequence is generally much longer than the target sequence. The probability of the target sequence is the sum of the probabilities of all corresponding intermediate paths:

\begin{gather}
  P_{ctc}(y|x)=\sum_{c\in\phi (y)}P(c|x)
\end{gather}
then, the CTC loss is formulated as:

\begin{gather}
  \mathcal{L}_{ctc}=-\sum_{(x,y)\in(\mathcal{X} ,\mathcal{Y}^{gloss})} \sum_{c\in\phi(y)} P(c|x)
\end{gather}
where \((\mathcal{X} ,\mathcal{Y}^{gloss})\) denotes the set of source video and target gloss pairs in the data.

We use cross entropy loss and length loss to optimize the auxiliary decoder and a kind of technology module, which can be formulated as:

\begin{gather}
  \mathcal{L}_{IF}=-\sum_{(x,y)\in(\mathcal{X} ,\mathcal{Y}^{gloss})}[log(P_{IF}(y|x))+(\tilde{S_x}-S_x  )^2]
\end{gather}
where \(S_x\) is the length of the target gloss, and \(\tilde{S_x}\) is the sum of all weight embedding \(W\), which represents the length of the prediction sequence.

We also use cross-entropy loss to optimize the main translation task of SimulSLT:

\begin{gather}
  \mathcal{L}_{hard}=-\sum_{(x,y)\in(\mathcal{X} ,\mathcal{Y})}log(P(y|x))
\end{gather}
where \((\mathcal{X} ,\mathcal{Y})\) denotes the set of video-text sequence pairs.

Therefore, the total loss function to train the SimulSLT model is:
\begin{gather}
  \mathcal{L} = \lambda_1\mathcal{L}_{ctc}+\lambda_2\mathcal{L}_{IF}+\lambda_3\mathcal{L}_{soft}+\lambda_4\mathcal{L}_{hard} 
\end{gather}
where the \(\lambda_1\) to \(\lambda_4\) are hyperparameters to trade off these losses, and the calculation method of \(\mathcal{L}_{soft}\) is the same as \(\mathcal{L}_{WORD-KD}\) proposed by \citet{kim-rush-2016-sequence}.

\begin{table*}[t]
  \centering
  \renewcommand\arraystretch{1.0}
  \setlength{\abovecaptionskip}{2pt}
  \setlength{\tabcolsep}{1.8mm}
    \caption{Comparisons of translation results on RWTH-PHOENIX-Weather 2014T dataset.}
    \label{table:main-result}
  \begin{tabular}{ccccccccccccc}
    \toprule
    \multirow{2}{*}{Method} & \multirow{2}{*}{k}& \multicolumn{5}{c}{DEV} & \multicolumn{5}{c}{TEST} \\
    \cmidrule(lr){3-7} 
    \cmidrule(lr){8-12}  
    &&BLEU-1         & BLEU-2 & BLEU-3 & BLEU-4 & ROUGE &BLEU-1 & BLEU-2 & BLEU-3 & BLEU-4 & ROUGE \\
    \midrule  
    Joint-SLT \cite{DBLP:conf/cvpr/CamgozKHB20} &inf&47.26 &34.40 &27.05 &22.38 & - &46.61 &33.73 &26.19 &21.32 & - \\
    SimulSLT(ours) &7 &{\bf 47.76} &{\bf 35.33} &{\bf 27.85} &{\bf 22.85} & 49.21 &{\bf 48.23}&{\bf 35.59}&{\bf 28.04}&{\bf 23.14} & 49.23\\
    \midrule
    SLT-Multitask \cite{Orbay2020} &inf &- &- &- &- & -&37.22&23.88&17.08&13.25&{\bf 36.28}  \\
    TSPNet-Joint \cite{DBLP:conf/nips/LiX0ZSSL20} &inf &- &- &- &-&-   &36.10 &23.12 &16.88 &13.41&34.96 \\
    SimulSLT(ours) &7 &36.21 &23.88 &17.41 &13.57&36.38   &{\bf 37.01} &{\bf 24.70} &{\bf 17.98} &{\bf 14.10}& 35.88 \\
    \midrule
    Conv2d-RNN(L) \cite{dataset3,DBLP:conf/emnlp/LuongPM15} &inf  &31.58 &18.98 &13.22 &10.00&32.60 &29.86 &17.52 &11.96 &9.00&30.70 \\
    Conv2d-RNN(B) \cite{dataset3,DBLP:journals/corr/BahdanauCB14} &inf  &31.87 &19.11 &13.16 &9.94&31.80 &32.24 &19.03 &12.83 &9.58&31.80\\
    SimulSLT(ours)&7  &{\bf 36.01} &{\bf 22.60} &{\bf 16.05} &{\bf 12.39}&{\bf 36.04 } &{\bf 35.92} &{\bf 22.70} &{\bf 16.03} &{\bf 12.27} & {\bf35.13 } \\
    \bottomrule
  \end{tabular}
 \end{table*}

\section{Experiments}
\subsection{Datasets}
We evaluate the SimulSLT model on the RWTH-PHOENIX-Weather 2014T (PHOENIX14T) dataset \cite{dataset3}, which is the most commonly used large-scale SLT dataset. Its data is collected from the weather forecast of the German public television station PHOENIX, including parallel sign language videos, gloss annotations, and corresponding text translations. We follow the official data set partitioning protocol, where the training set, validation set, and test set contain 7096, 519, and 642 samples, respectively. The dataset contains consecutive sign language videos from 9 different signers with a vocabulary of 1066 different signs. The text annotations in the dataset are German spoken language with a vocabulary of 2887 different words.

\subsection{Implementation Details}
Our model's main structure is the Transformer \cite{DBLP:conf/nips/VaswaniSPUJGKP17}, which has been widely used in machine translation. The number of hidden units in the model, the number of heads, the number of encoder and decoder layers are set to 512, 8, 3, 3, respectively. We also use dropout with 0.3 and 0.6 drop rates on encoder and decoder layers to mitigate overfitting. We use the word embeddings trained by FastText \cite{bojanowski2016enriching} on the Wikipedia German dataset as the initial word embeddings of the model, and the word embeddings of gloss are initialized randomly. We adopt Xavier initialization \cite{DBLP:journals/jmlr/GlorotB10} to initialize our network. 


\subsection{Training Setup}
We train our model on a single Nvidia 2080ti GPU with a total batch size of 32. We use the Adam optimizer with a learning rate of \(5 \times 10^{-4}\) (\(\beta_1 \)=0.9, \(\beta_2\)=0.998), and the weight decays to \(10^{-3}\). We utilize the plateau learning rate schedule to update the learning rate, which tracks the BLEU \cite{DBLP:conf/acl/PapineniRWZ02} score on the validation set, the patience and factor are set to 9 and 0.5, respectively. The validation set is evaluated every 100 steps. During validation, we use a beam search algorithm with a beam size of 3 and a length penalty value -1 to decode the text sequence. Training ends when the learning rate is less than \(10^{-7}\). During training, the weights of the loss function \(\lambda_1 - \lambda_4\) are set to 10, 1, 0.6, and 0.4 respectively. 

\subsection{Evaluation Metrics}
We use the BLEU \cite{DBLP:conf/acl/PapineniRWZ02} and ROUGE-L \cite{doddington2002automatic} scores to evaluate the quality of the translation. We use average lagging (AL) \cite{DBLP:conf/ijcnlp/MaPK20} and average proportion (AP) \cite{DBLP:journals/corr/ChoE16} to evaluate the latency of our model. The former is used to measure the degree of out of sync between the model output and the signer, and the latter is used to measure the average absolute delay cost of each target text. Assuming that the input video sequence of the model is \(X=(x_1,x_2,...,x_n)\), the reference target text sequence is \(Y^*=(y_1^*,y_2^*,...,y_n^*)\), and the predicted target text sequence is \(Y=(y_1,y_2,...,y_n)\). The calculation formula of AL is as follows :
\begin{gather}
  \mathrm{AL}=\frac{1}{\tau(|\boldsymbol{X}|)} \sum_{i=1}^{\tau(|\boldsymbol{X}|)} d\left(y_{i}\right)-\frac{|\boldsymbol{X}|}{\left|\boldsymbol{Y}^{*}\right|} \cdot T_{s} \cdot(i-1)
\end{gather}
where \(T_{s}\) represents the time interval between obtaining two video frames, \(|\boldsymbol{Y}^*|\) is the length of the reference translation text, \(|\boldsymbol{X}|\) is the length of the input video, and \(\tau(|\boldsymbol{X}|)\) is the index of the first target token generated when the model has read all input videos. \(d(y_i)\) represents the time it takes to generate \(y_i\). \(d(y_i)\) is equal to \(T_{s}\) multiplied by the number of video frames that has been read when \(y_i\) is generated.
The calculation formula of AP is as follows:
\begin{gather}
  AP=\frac{1}{T_s\cdot|X|\cdot|Y|}\sum_{i=1}^{|Y|}t(i)
\end{gather}
where, \(t(i)\) represents the duration of the model from the beginning of the calculation to the generation of the \(i\)-th target text.

\subsection{Experimental Results}
\subsubsection{Translation Accuracy}
In this section, we compare the performance of SimulSLT with the existing end-to-end non-simultaneous sign language translation model and explore the impact of different k settings on the model performance. The comparison results are shown in Table \ref{table:main-result}. In order to provide a broad baseline for future research in this field, we also tested the SimulSLT model trained without gloss annotations. In this case, we simply use \(p = mean(\frac{v_i}{g_i})\) to segment the video statically, where \(v_i\) represents the length of the video and \(g_i\) represents the length of the gloss. In addition, the native modal in Table \ref{table:ab} also uses this video segmentation method.

According to the different data types used when training the model, we divided the comparison results into three groups. All models of the first group are trained with all the data set information, including gloss annotations (here we use the same CNN network as \cite{DBLP:conf/cvpr/CamgozKHB20} as the feature extractor.) Although the models in the second group did not use gloss during training, they all used other additional data related to sign language. For example, TSPNet-Joint \cite{DBLP:conf/nips/LiX0ZSSL20} uses the additional American Sign Language video and its corresponding gloss information provided in the two data sets \cite{li2020word,DBLP:conf/bmvc/JozeK19}, and both SLT-Multitask \cite{Orbay2020} and SimulSLT use the additional gesture information provided in the dataset \cite{DBLP:conf/cvpr/KollerNB16}. We use the pre-trained CNN network in \cite{DBLP:conf/cvpr/KollerNB16} to extract hand shape features as the input of the model. All models in the third group are not trained with gloss annotations, and the CNN network (we use EfficientNet \cite{DBLP:conf/icml/TanL19}) pre-trained on ImageNet \cite{DBLP:conf/cvpr/DengDSLL009} is used as the feature extractor. We did not list the model in \cite{Yin2020} in the table because it is a two-stage model of sign2gloss2text. As shown in the table, SimulSLT with k set to 7 achieves the best performance compared to other non-simultaneous sign language translation models in the three groups. These results show that our model still has excellent performance under the condition of lower latency.

Table \ref{table:dff-k} shows the BLEU scores of SimulSLT trained with different k. We can sacrifice a part of the performance to obtain lower latency by setting a smaller k.

\begin{table}[htb]
  \centering
  \setlength{\abovecaptionskip}{2pt}
  \caption{The BLEU scores of SimulSLT trained with different data on the test dataset of PHOENIX14T.}
  \begin{tabular}{lcccc}
    \toprule  
    model&k=1&k=3&k=5&k=7\\
    \midrule  
    SimulSLT(with gloss)     &19.27&21.62&22.58&23.14\\
    SimulSLT(additional data)&12.66&12.83&13.46&14.10\\
    SimulSLT(without gloss)  &11.24&11.37&11.91&12.27\\
    \bottomrule 
    \end{tabular}
    \label{table:dff-k}
\end{table}

\subsubsection{Translation Delay}
\begin{figure}[htbp]
  \setlength{\abovecaptionskip}{2pt}
  \centering
  \subfigure[The translation quality against the latency in terms of AL.]{
    \begin{minipage}[t]{\linewidth}
      \centering
      \includegraphics[width=\linewidth]{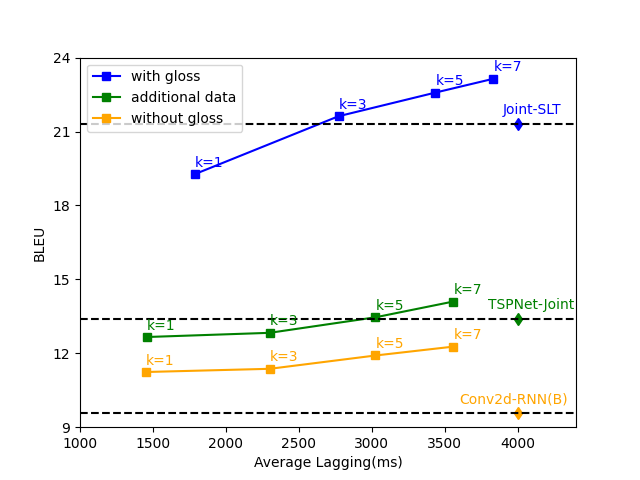}
      \label{fig:nal}
    \end{minipage}
  }
  \vspace{-3mm}
  \\
  \subfigure[The translation quality against the latency in terms of AP.]{
    \begin{minipage}[t]{\linewidth}
      \centering
      \includegraphics[width=\linewidth]{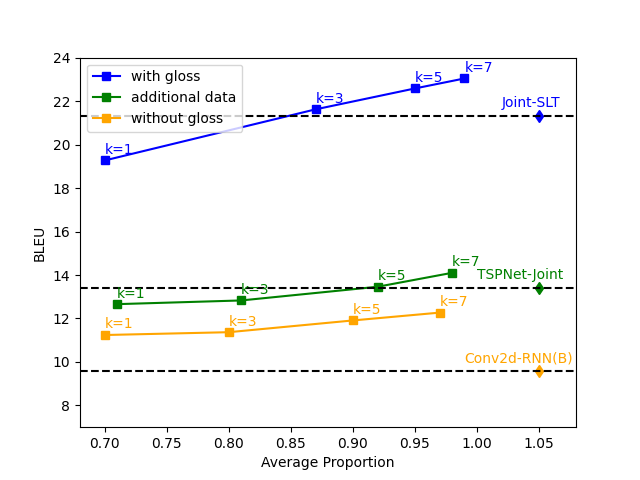}
    \end{minipage}
  }
  \centering
  \caption{The translation quality against the latency metrics (AL and AP) on PHOENIX14T dataset.}
\end{figure}

In this section, we explore the relationship between translation delay and translation quality, and compare our model with the most advanced end-to-end non-simultaneous sign language translation model. We plot the translation quality (denoted in BLEU scores) against the latency metrics (AL) of SimulSLT models trained on different data in Figure \ref{fig:nal}. We can see that the performance of the model gradually improves as k becomes larger, but the translation delay is gradually becoming larger. Therefore, in actual application, we can set k according to project requirements to achieve a balance between quality and delay.

The dotted line in the figure shows the comparison between our model and the non-simultaneous model. We can find that SimulSLT surpasses the non-simultaneous model in terms of translation performance and translation delay. Especially when no gloss is used, even when k is set to 1, SimulSLT is higher than the most advanced non-simultaneous model. In terms of latency, SimulSLT can achieve the same performance as the non-simultaneous model in almost half the time.

\subsection{Ablation Study}
In this subsection, we will introduce the results of our ablation experiments on the PHOENIX14T dataset, and analyze the effectiveness of our proposed method through the experimental results. The experimental results in the case of using gloss annotations are shown in Table \ref{table:ab}. We can find that our proposed method can improve the translation accuracy of different wait-k.

\begin{table}[htb]
  \centering
  \setlength{\abovecaptionskip}{2pt}
  \setlength{\tabcolsep}{3mm}
  \caption{Results of ablation experiments on the PHOENIX14T dataset. Native SLT represents a model that uses a fixed-length segmentation video and then applies a wait-k strategy. We gradually add the methods we mentioned earlier to verify their effectiveness.}
  \begin{tabular}{lccc}
    \toprule  
    Model&k=1&k=3&k=7\\
    \midrule  
    Native SLT     &10.07&14.12&17.48\\
    \midrule
    +BP             &18.34&19.99&20.65\\
    +BP+CTC         &18.75&20.42&22.32\\
    +BP+KD          &18.88&20.48&22.68\\
    +BP+Re-encode   &18.81&20.74&22.45\\
    \midrule
    +BP+Aux+Re-encode+KD \\(SimulSLT) &{\bf 19.27 }&{\bf 21.62}&{\bf 23.14}\\
    \bottomrule 
    \end{tabular}
    \label{table:ab}
\end{table}
\subsubsection{The Effectiveness of Boundary Predictor}
As shown in Table \ref{table:ab}, since the video segment corresponding to a gloss in sign language is not of a fixed length, simply using a fixed segmentation will cause information misalignment and insufficient information obtained during decoding. By adding a boundary predictor, the model can better learn the correspondence between the video and the gloss, and generate a more accurate boundary. From the BLEU score in the second row, it can be seen that the translation accuracy with different wait-k can be improved by adding a boundary predictor to Native SLT model.

\subsubsection{The Effectiveness of CTC}
The result in the third row shows that the translation accuracy of the model can be further improved by adding an auxiliary connectionist temporal classification (CTC) decoder. Through the supervision of CTC loss, the encoder can have more robust feature expression ability, and the boundary predictor can also learn the alignment information better. 

\begin{figure}[htbt]
  \setlength{\abovecaptionskip}{2pt}
  \includegraphics[width=\linewidth]{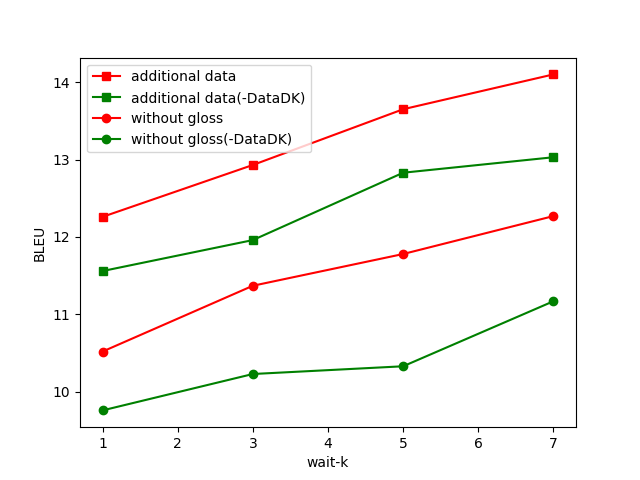}
  \caption{Experimental results of the impact of knowledge distillation on the translation accuracy of models trained with additional data and models trained without gloss. The green line represents the BLEU score without knowledge distillation, and the red represents the BLEU score after use.}
  \label{fig:dk}
\end{figure}

\subsubsection{The Effectiveness of Knowledge Distillation}
We further explored the impact of knowledge distillation on model performance(show Row 4 vs Row 2). Knowledge distillation can transfer the knowledge learned by the teacher model to the student model, which reduces the difficulty of optimizing the student model. The results show that knowledge distillation at the data level makes the model achieve a greater performance improvement. We also tested the impact of knowledge distillation on the performance of the model in the other two cases, and the results are shown in Figure \ref{fig:dk}. It can be seen that by adding knowledge distillation, the model has achieved an improvement in the translation accuracy in the different wait-k settings in the two cases.

\begin{table}[htb]
  \centering
  \setlength{\abovecaptionskip}{2pt}
  \setlength{\tabcolsep}{1.5mm}
  \caption{Experimental results on the PHOENIX14T dataset using the re-encode method and the original method.}
  \begin{tabular}{lccccc}
    \toprule  
    Data&Model&k=1&k=3&k=7\\
    \midrule  
    additional data&SimulSLT(-Re-encode) &11.94&12.00&12.71\\
    additional data&SimulSLT&12.66&12.83&14.10\\

    \toprule
    without gloss&SimulSLT(-Re-encode)  &10.23&10.61&11.73\\
    without gloss&SimulSLT&11.24&11.37&12.27\\
    \bottomrule 
    \end{tabular}
    \label{table:re1}
\end{table}

\subsubsection{The Effectiveness of Re-encode}
We further explored the impact of the re-encode method on the performance of the model. In the first step, we conducted experiments to analyze whether the re-encode method shown in Figure \ref{fig:re-encode2} can improve the translation accuracy of the model compared to the original method shown in Figure \ref{fig:re-encode1}. The experimental results are shown in Table \ref{table:ab} (Row 5 vs Row 2) and Table \ref{table:re1}. It can be seen that in all cases, compared with the original method, the re-encode method can help the model improve the translation accuracy. 

\begin{table}[htb]
  \newcommand{\tabincell}[2]{\begin{tabular}{@{}#1@{}}#2\end{tabular}}
  \centering
  \setlength{\abovecaptionskip}{2pt}
  \caption{Ablation study of different fusion methods.}
  \begin{tabular}{lccc}
    \toprule  
    Type&\tabincell{c}{with gloss\\(k=7)}&\tabincell{c}{additional data\\(k=7)}&\tabincell{c}{withou gloss\\(k=7)}\\
    \midrule  
    Re-encode-only&22.98&13.89&12.01\\
    concat&22.89&13.75&11.89\\
    add&23.14&14.10&12.27\\
    \bottomrule 
    \end{tabular}
    \label{table:fusion}
\end{table}

\subsubsection{The impact of different fusion methods}
We further explored the impact of different fusion methods as shown in Figure \ref{fig:main-b} on the performance of the model. The experimental results are shown in Table \ref{table:fusion}. It can be found that the best translation accuracy rate is obtained by using the direct addition fusion method.

\begin{figure}[tbp]
  \setlength{\abovecaptionskip}{2pt}
  \includegraphics[width=\linewidth]{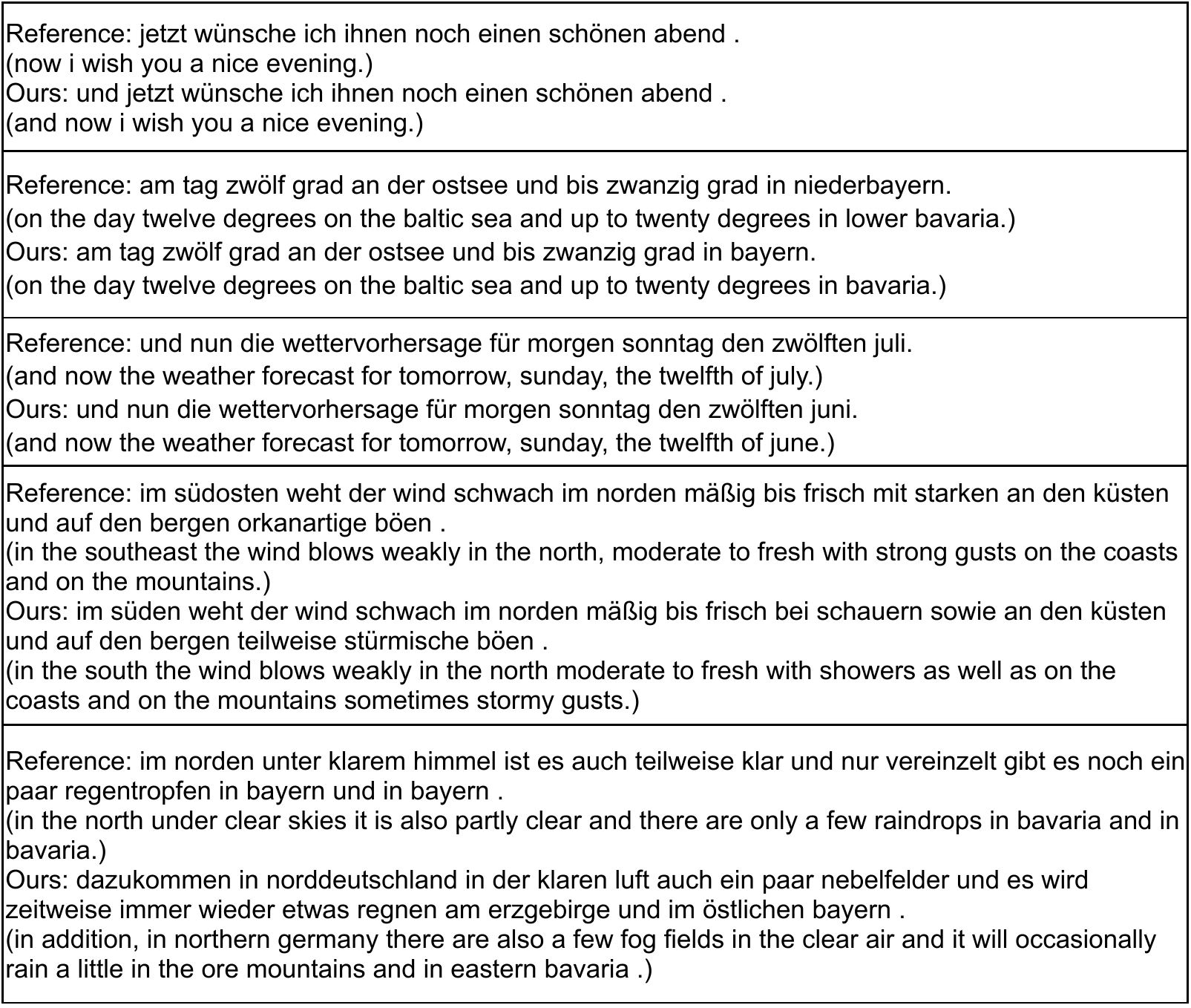}
  \caption{Comparison of the spoken language generated by the SimulSLT model and the reference text.}
  \label{fig:rp}
\end{figure}

\subsection{Qualitative Results}
In this section we show some qualitative results of our model. As shown in the first and second row in Figure \ref{fig:rp}, our model performs well when translating short sentences, but there is still some difference with the ground truth when translating long sentences (show in Row 3 and Row 4). The front part of the long sentence predicted by the model is quite different from the true value. As mentioned in \cite{dataset3}, the order of sign language video and spoken language is not the same, and the sign language video corresponding to the former part of spoken language semantics may be in the latter part. Therefore, on the one hand, the model does not get the corresponding semantic information when translating the front part of the long sentence; on the other hand, long sentences pose a greater challenge to the model’s context prediction ability. These two reasons lead to this problem. In the case of lenient latency requirements, we can use larger k to alleviate this problem. In addition, we can also alleviate this problem by building larger datasets and models with stronger contextual predictive capabilities.

\section{Conclution}
In this work, we developed SimulSLT, an end-to-end simultaneous sign language translation system, which can continuously translate sign language videos into spoken language. To achieve controllable low-latency translation, we introduce a wait-k strategy to control the read and write of the model. We divide the continuous sign language video into discrete video segments by introducing an a kind of technology based boundary predictor, which is inspired by the way human neurons work in biology. To bridge the accuracy gap between the simultaneous sign language translation model and non-simultaneous sign language translation, we propose a novel re-encode method to help the model obtain more abundant contextual information, which allows the video features to interact more fully. Knowledge distillation is also used to transfer knowledge from the non-simultaneous sign language translation teacher model to the simultaneous sign language translation student model, which further improves the performance of the model and reduces the optimization difficulty of the model. In order to better help the boundary predictor learn alignment information and enhance the feature extraction capability of the encoder, we propose a series of methods including auxiliary gloss decoding task and CTC auxiliary decoding. Experiments on the PHOENIX14T dataset show that SimulSLT can achieve a translation accuracy that exceeds the latest end-to-end non-simultaneous sign language translation model while maintaining a much lower latency.

\section*{Acknowledgments}
This work was supported in part by the National Key R\&D Program of China (Grant No. 2018AAA0100603), National Natural Science Foundation of China under Grant No.61836002, No.62072397 and Zhejiang Natural Science Foundation under Grant LR19F020006. This work was also partially supported by the Huawei Noah's Ark Lab.

\newpage

\bibliographystyle{ACM-Reference-Format}
\balance
\bibliography{draft}
\end{document}